IAC–24–B3,7,12,x87709

# TOWARDS A RELIABLE OFFLINE PERSONAL AI ASSISTANT FOR LONG DURATION SPACEFLIGHT

Oliver Bensch[a*], Leonie Bensch[b], Tommy Nilsson[c], Florian Saling[d], Wafa M. Sadri[e], Carsten Hartmann[f], Tobias Hecking[g], J. Nathan Kutz[h]

[a] *German Aerospace Center (DLR), Cologne, Germany, oliver.bensch@dlr.de*
[b] *German Aerospace Center (DLR), Cologne, Germany, leonie.bensch@dlr.de*
[c] *European Space Agency (ESA), Cologne, Germany, tommy.nilsson@esa.int*
[d] *German Aerospace Center (DLR), Cologne, Germany, florian.saling@ext.esa.int*
[e] *German Aerospace Center (DLR), Cologne, Germany, wafa.sadri@ext.esa.int*
[f] *German Aerospace Center (DLR), Wessling, Germany, carsten.hartmann@dlr.de*
[g] *German Aerospace Center (DLR), Cologne, Germany, tobias.hecking@dlr.de*
[h] *Department of Applied Mathematics, University of Washington, Seattle, WA USA, kutz@uw.edu*
[*] *Corresponding author*

## Abstract

As humanity prepares for new missions to the Moon and Mars, astronauts will need to operate with greater autonomy, given the communication delays that make real-time support from Earth difficult. For instance, messages between Mars and Earth can take up to 24 minutes, making quick responses impossible. This limitation poses a challenge for astronauts who must rely on in-situ tools to access the large volume of data from spacecraft sensors, rovers, and satellites, data that is often fragmented and difficult to use.

To bridge this gap, systems like the Mars Exploration Telemetry-Driven Information System (METIS) are being developed. METIS is an AI assistant designed to handle routine tasks, monitor spacecraft systems, and detect anomalies, all while reducing the reliance on mission control. Current Generative Pretrained Transformer (GPT) Models, while powerful, struggle in safety-critical environments. They can generate plausible but incorrect responses, a phenomenon known as "hallucination," which could endanger astronauts.

To overcome these limitations, this paper proposes enhancing systems like METIS by integrating GPTs, Retrieval-Augmented Generation (RAG), Knowledge Graphs (KGs), and Augmented Reality (AR). The idea is to allow astronauts to interact with their data more intuitively, using natural language queries and visualizing real-time information through AR. KGs will be used to easily access live telemetry and multimodal data, ensuring that astronauts have the right information at the right time.

By combining AI, KGs, and AR, this new system will empower astronauts to work more autonomously, safely, and efficiently during future space missions.

## 1. Introduction

As humankind prepares for a new generation of crewed missions to the Moon and Mars, these endeavors introduce a unique set of challenges. One of the most significant is the limitation in communication due to bandwidth and latency, which prevents real-time interaction between astronauts and mission control. For example, communication delays between Mars and Earth take up to 24 minutes, depending on planetary alignment. Similarly, the Artemis program's focus on the lunar south pole will make radio blackouts a recurring challenge for future expeditions to the Moon [1]. These limitations will require future astronauts to operate with less direct guidance and oversight from terrestrial control centers. Instead, they will need to exercise greater autonomy whilst relying on advanced in-situ tools for real-time guidance and other support. Despite the amount of data available from, for example, spacecraft sensors, rovers, and satellite systems, this multimodal information is often fragmented across diverse formats. For astronauts to benefit from it, the data must be processed, integrated, and presented in an easily accessible manner.

AI-based systems, described in this paper as Intelligent Personal Assistants (IPAs), such as the Crew Interactive Mobile Companion (CIMON) deployed on the International Space Station (ISS), have illustrated the potential of AI to assist astronauts. However, these systems are constrained by their reliance on predefined, inflexible responses. Furthermore, they are limited to specific data sources, such as structured procedures in formats like the Operational Data File (ODF), which contains step-by-step instructions used by astronauts to perform various tasks, including maintenance, experiments, and system operations. These procedures ensure consistency, safety, and accuracy in task completion within the space environment. They are essential for routine operations and managing emergencies.

To overcome these limitations, the Mars Exploration Telemetry-Driven Information System (METIS) is currently being developed, with certain components already tested using real data from the Columbus module onboard the International Space Station (ISS) [2]. METIS is an AI assistant designed to enhance astronaut autonomy by managing routine operations, executing tasks, monitoring spacecraft systems, detecting anomalies, and aiding in mission planning. The primary objective of METIS is to reduce reliance on Earth-based mission control during space operations. METIS employs a multi-agent system (MAS) architecture comprising four key agents: the monitoring agent, reasoning agent, planning agent, and commanding agent. The monitoring agent continuously tracks telemetry data and triggers alerts to the reasoning agent when anomalies are detected. The reasoning agent, which utilizes a Case-Based Reasoning (CBR) framework, assesses the situation by referencing historical cases and generates an appropriate response. If a suitable solution is identified, it is handed off to the planning agent, which adjusts the mission timeline accordingly. If no solution is found, the issue is escalated to ground operators. Finally, the commanding agent executes the revised mission plan [3].

While METIS has outlined a way to reduce the reliance on ground support during long-duration space missions, some research challenges remain. A key focus is the integration of unstructured data, such as operational manuals, and multimodal data, including images, as well as the development of an intuitive interface that allows astronauts to interact with the system through natural language and visualizes the information accordingly. It should also be mentioned here, that the main goal of METIS is to be able to perform all current ground activities autonomously on-board. It is therefore essential that the system is both able to execute system operations (e.g., system maintenance activities) with minimal crew interaction, while also providing adequate assistance to crew whenever needed and/or appropriate (e.g., during on-board anomalies).

To address these challenges, this paper proposes extending the capabilities of systems, such as METIS, by incorporating Generative Pretrained Transformers (GPTs), Knowledge Graphs (KGs), and Augmented Reality (AR) in one IPA. These enhancements aim to improve astronaut support during long-duration missions by integrating data from a wide array of sources, including satellites and rovers, while leveraging the advanced reasoning capabilities exhibited by GPT models across various applications. Most importantly, this extension enables astronauts to access crucial real-time information through natural language and visualize the corresponding data seamlessly.

## 2. AI-Based Personal Assistants in Space Exploration

*2.1 Limitations of GPT-based assistants*

IPAs that would rely solely on GPTs are limited by the quality and scope of their training data. For instance, models such as GPT-4 [4], or offline deployable alternatives like LLaMA 3.1 [5] or Mixtral8x7B [6], depend heavily on the data they were trained on.

A key limitation of these models for safety-critical situations are "hallucinations" [7]. Hallucinations occur when the model generates information that appears plausible but is not based on the underlying data it was trained on. GPT-based models generate responses by predicting what word or phrase is statistically likely to follow the previous input. When asked about topics outside their training data or when sufficient data is not available, the model may fabricate information to complete its response. While this output can often seem coherent, it may include inaccuracies or outright falsehoods, which is particularly problematic in safety-critical contexts where reliable and verifiable information is crucial [8].

For example, during an Extravehicular Activity (EVA, commonly known as "spacewalk"), an astronaut might ask the IPA for specific instructions on fixing a component on the ISS. If the assistant's training data does not include the latest EVA procedures or detailed technical manuals for that specific component, the model may fabricate plausible-sounding steps. It might recommend using a tool that is not suitable or advise the astronaut to perform an action that could lead to equipment damage or endanger the crew's safety. Since the generated response is presented with the same confidence as accurate information, the astronaut would have no easy way to distinguish between correct instructions and hallucinated, potentially dangerous advice.

Although fine-tuning the model on domain-specific data, such as predefined procedures, can reduce the likelihood of hallucinations, it does not eliminate the problem entirely. Fine-tuned models are still constrained by their training data and cannot integrate real-time, multimodal data (such as live telemetry or sensor data from the ISS) into their reasoning. Furthermore, even after fine-tuning, the generated answers cannot be verified by the user, which poses an ongoing risk in high-stakes environments like space exploration. In domains where safety is critical, such as aerospace, the inability to fully trust the model's responses remains a significant concern [9].

*2.2 Retrieval-Augmented Generation (RAG)*

Retrieval-Augmented Generation (RAG) [10] combines GPT models with a document collection by indexing the collection via model embeddings into a vector database. RAG first generates an embedding of the user's query and matches it against the vector database to retrieve relevant documents. These documents are then combined with the user's query to generate an answer using the GPT model. This approach enhances the accuracy of the responses and ensures that the model is not restricted to the information in its training set. Additionally, the documents used to generate an answer can be linked via the user interface, allowing users to verify the output. Techniques such as sentence-based embeddings or semantic chunking can further improve the accuracy of identifying relevant text segments in longer documents, even when the user query is brief [11]. For instance, an astronaut facing an issue with a tool on the Moon could query an IPA based on a GPT model using RAG, which retrieves relevant text blocks from a multi-page tool manual and then provides the astronaut with a concise summary and the corresponding sections from the documentation.

However, astronauts may need to access information in other formats, such as telemetry data, which is typically structured, like time-series data, rather than presented in text blocks. Embedding-based approaches often struggle when handling structured data that includes numerical information, such as dates or temperatures. This is because vector embeddings are designed to capture semantic relationships in textual data but often fail to accurately represent numerical values without special adjustments [12]. As a result, numerical data may be mapped to similar points in vector space, regardless of their actual differences, leading to potential inaccuracies in retrieval [10]. Moreover, an RAG-based approach can only consider indexed data within the vector database and is therefore unable to access live telemetry.

To address these limitations, techniques such as SQL query generation or hybrid search [11], where traditional search methods are combined with RAG, can be employed. Nevertheless, these systems do not take multimodal data, such as images or video, into account or struggle when required to combine information from multiple documents [11].

Procedures in ODF format, commonly used in the space domain, might include various types of data, such as images, videos, audio, or even 3D files. Additionally, information in these procedures is frequently presented as bullet points rather than continuous text and may contain references to other procedures.

Indexing these types of data can be challenging for standard RAG-based or hybrid search approaches. User queries regarding procedural steps that reference multiple documents, or that involve multimodal data, may not be properly identified by an RAG-based assistant.

## 3. Integration of Knowledge Graphs and GPT Models

*3.1 Data Pre-Processing Using (Multimodal) Large Language Models*

METIS has demonstrated how structured and semi-structured data can be effectively utilized to monitor telemetry and detect anomalies through its monitoring agent [13]. However, the system can also benefit from unstructured textual data sources, such as on-board system manuals, which contain important information, such as emergency scenarios.

Natural language processing techniques, including keyword extraction, taxonomy tagging [14], and named entity recognition (NER) [15], can be employed to automatically identify, classify, and structure this relevant textual information. Recent research has shown that tools like GraphRAG [16] even enable the automatic conversion of large unstructured text corpora into KGs.

As stated in section 1, it is also essential to account for multimodal data, such as images captured by rovers or satellites. Recent work has outlined a first step towards how RAG-based approaches can be used for ISS procedures, while also considering referenced multimodal data using KGs and visualize that information using AR [17].

However, further data sources such as telemetry or images from rovers and satellites might also be of interest for integration into a multi-agent system like METIS. Pipelines leveraging the latest multimodal large language models (MLLMs), such as Pixtral 12B [18], can be used to automatically process and enrich such data sources, including images or rendering of 3D-files, by adding textual metadata.

Nevertheless, classical RAG or hybrid search systems might struggle to identify requested images from user questions by short annotated meta information.

*3.2 Knowledge Graphs to Combine Multimodal Data for Space Operations*

KGs are structured representations, where nodes represent information points and edges depict the relationships between them. Many space exploration procedures are stored in the ODF format, which includes references to other relevant procedures, making them well-suited for interpretation as a graph structure. As described in section 3.1, this KG can be further interlinked using NLP methods and combined with other unstructured textual information, such as manuals.

Furthermore, the nodes and edges in such knowledge graphs can represent various types of data, allowing them to incorporate multimodal information, including live images or telemetry from rovers and satellites. As described in section 3.1, MLLMs can be utilized to further incorporate multimodal data in the graph by adding textual meta information.

Using this approach, a METIS reasoning agent, for instance, could access and be trained on a broader range of data sources, including unstructured and multimodal data.

*3.3 Language Models to Support Ground-Control and Crew Members*

Language models can not only be used to process and structure multimodal data into a KG. The latest GPT models have demonstrated potential in detecting mental health conditions [19] or even providing support to individuals experiencing psychological difficulties [20].

Furthermore, the style of interaction, such as short or long responses or generating answers in a friendly or strict manner, can be adjusted either via prompting or by activation engineering [21], depending on requirements. Here the IPA could change its response style, for instance, to be similar to a human colleague or friend.

Moreover, as described in section 2.2, chat systems based on RAG or GraphRAG could constitute a viable approach to making unstructured textual information, such as manuals of on-board systems, accessible in natural language for crew members during long-duration space flights. By integrating KGs with GPT models', astronauts can query the KG using natural language, and the AI retrieves accurate information from the KG. Utilizing the power of graphs, multimodal data, such as live telemetry, can thus be taken into account by such a system.

GPT models reasoning capabilities, as shown, for example, in OpenAI's recent o1 evaluation [22], could not only support the reasoning or planning agent in a MAS architecture, as described in METIS, but also support crew or ground control in the solution-finding process. While classic RAG systems would only identify and consider matching documents based on their shared embedding in the reasoning process, such a system could access additional related important documents during reasoning, based on their relations in the KG. Not only the generated answer but also the reasoning process, the utilized text segments of documents, and the identified related items in the KG could therefore be shared with the crew or operators.

# 4. Head-Mounted and Head-Up Display Integration

While the principles outlined above enable AI-based personal assistants to generate relevant information that supports astronauts' daily tasks, a key challenge lies in seamlessly integrating this information into their workflows. With mission procedures typically requiring extensive manual work, traditional handheld devices, such as tablets, may prove impractical. A promising alternative involves the use of Head-Up displays (HUDs) or Head-mounted or Helmet-mounted displays (HMDs) that overlay computer-generated information directly onto the astronaut's field of view, thereby enhancing both the intuitiveness and accessibility of the generated information.

*4.1 Displaying 2D Information in the Field of View Using Head-Up Displays*

Early research into HUDs and HMDs for human spaceflight includes NASA's work on head-mounted displays in the 1980s [23, 24]. Although their use was originally deferred due to high costs, advancements in display technologies over the following decades have led to a resurgence of interest.

Today, private space companies such as ILC Dover [25] are actively developing future astronaut suit models with integrated HUD technology. Recently, SpaceX announced that their EVA suit will feature a 3D-printed helmet with a camera and a HUD integrated into the visor, capable of tracking the suit's internal conditions. This system was first launched during the Polaris Dawn mission in September 2024 [26].

Research has explored 2D HUD features that display critical information, such as heart rate and elapsed EVA time, allowing astronauts to monitor their physical condition independently of ground control [27]. Similarly, the integration of HUDs with speech-recognition systems has been proposed to enable astronauts to access biometric data during EVAs, enhancing both their safety and autonomy [28].

In this context, IPAs could support the interpretation and modification of the information displayed on the 2D HUD. Astronauts could access and modify the content of the HUD, switching between alternative information, such as checklist steps, suit diagnostics, biometrics, environmental data, or EVA time retrieved from the KG using natural language.

Switching between different HUD data can be made more intuitive by using natural language queries and suggesting linked content, while the HUD screen occupies minimal space inside the astronaut's visor. This design minimizes obstruction of the astronaut's field of view, which is crucial for maintaining situational awareness during EVAs [29].

Combining visual information with audible warnings provided by the IPA could further reduce visual clutter and enhance the communication of safety-critical information, especially in a high workload situation [30].

Sound integration in HUD systems also offers significant potential for improving astronaut performance during lunar missions. In other domains, auditory cues have been shown to support navigation and alert users to critical information [31]. For deep space missions, audio cues could confirm selections, draw attention to urgent messages, or augment visual data during navigation tasks. Previous research in automotive and aviation industries demonstrates the effectiveness of auditory-visual integration for hazard warnings and situational awareness [32].

*4.2 Displaying Spatial Augmented Reality Using Head-Mounted Displays*

Rather than simply overlaying the user's view with non-spatial 2D information using HUDs, HMDs can also project 3D visualizations directly onto the real-world environment, while the content is continuously adjusted based on where the user is looking. AR cues can help visualize complex information in an intuitive manner.

By reducing visual clutter and enhancing users' mental models, AR cues improve task performance and reduce cognitive load [33, 34].

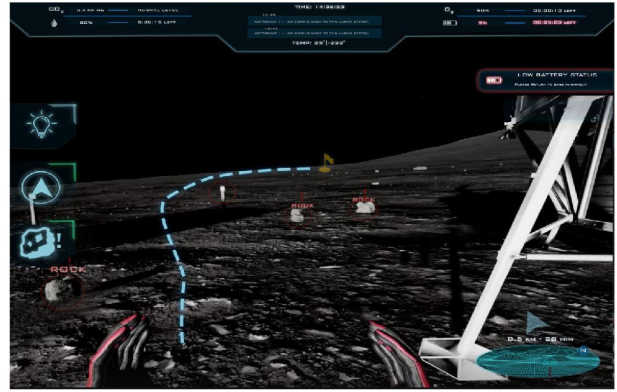

Fig. 1. A conceptual prototype illustrating potential content displayed using a HMD or HUD for astronauts during lunar surface operations.

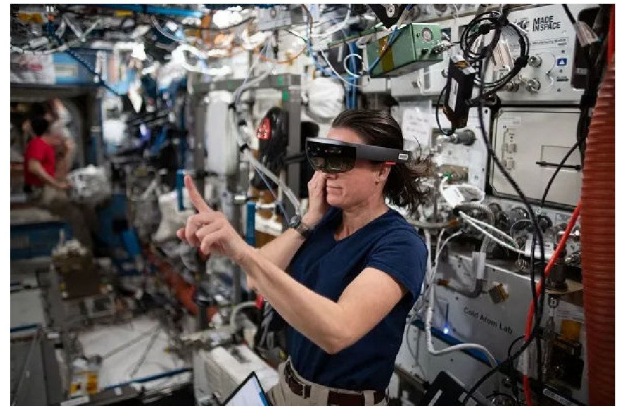

Fig. 2. Astronaut Megan McArthur uses a HoloLens HMD onboard ISS. Credits: NASA.

Integrating 3D-generated content into the astronaut's field of view also reduces their need to look down at a display or interpret written information, as is typical in current procedures.

Drawing on such advantages, AR cues have demonstrated the ability to improve situational awareness, reduce workload, and improve performance in various contexts, including routine simulated procedures aboard the ISS [35, 36] and simulated lunar navigation tasks [29].

When combined with AR cues, IPAs can further reduce cognitive workload by providing astronauts with easy access to enhanced spatial navigation tools, terrain awareness features, as well as intuitive maintenance and repair instructions.

*4.2.1 Navigation*

With the new generation of lunar landings set to begin in the near future, navigating the inhospitable surface of the Moon has been predicted to pose significant safety challenges. Particularly so in areas with extreme lighting conditions, such as the dark shadows and bright sunlight found at the lunar poles, increasing the risk of astronauts tripping on rocks or other obstacles. Moreover, the lack of atmospheric falloff and the absence of vertical features makes distance estimation difficult.

Emerging research suggests that the use of integrated AR navigation systems could play a key role in mitigating some of these risks. For example, Bensch et al. [29] demonstrated that AR cues could be advantageous for astronauts navigating complex lunar terrains, enhancing safety and efficiency while reducing cognitive load [37]. For example, warnings concerning surrounding obstacles and tripping hazards, along with spatial safety zones, common in robotics applications, could overlay the astronaut's field of view to provide critical guidance in real-time [38].

Bensch et al. [29] also showed that astronauts prefer the use of conformal AR features, meaning visual elements that seamlessly integrate into the environment. Waypoints placed in real-world locations, for instance, are preferable over non-spatial navigational elements. Both the generated answer and the reasoning process, including the utilized text segments of documents and the identified related items in the KG, could therefore be shared with the crew or operators.

In this context, IPAs could be used to dynamically update navigational cues based on new knowledge. For example, IPAs could analyze sensor data, such as LIDAR or camera information in real time and adjust routes or warn astronauts of upcoming navigation changes based on mission updates. This would allow astronauts to focus on immediate tasks while the IPA anticipates potential hazards or changes in terrain conditions.

Given the advantages of KGs and RAG, fitting information could thus be found reliably and be intuitively added to the real-world content. Furthermore, it could also be accessed in the required level of detail by the astronaut.

*4.2.2 Terrain Awareness*

In combination with adding relevant navigational information, terrain awareness features could greatly reduce the risks posed by hazardous terrain and poor visibility, especially during lunar or Martian expeditions.

Here, for instance, the use of 3D grid meshes projected onto real-world environments has been shown to improve the user's understanding of terrain characteristics and distances [39]. Moreover, the system could project spatial safety zones and obstacle warnings (such as rocks, equipment, or moving objects) on the environment through the HMD. The IPA could deliver audio notifications when astronauts approach these areas, drawing attention to imminent risks and could answer queries related to specific hazards while updating linked navigational information visually in the HMD.

Although night vision tools are not able to work in space conditions, similar concepts, e.g., based on LIDAR or other sensor data, could be utilized to enhance the vision of astronauts during EVAs.

Also, incorporating real-time data from orbiters with geological composition maps and terrain awareness systems could significantly improve how astronauts engage with their environment. By overlaying geological information, identifying areas of interest, and providing specific sampling instructions, the combination of IPA technology and AR integration could improve the understanding, awareness, and interaction with the surrounding terrain [40].

*4.2.3 Instructions*

AR technology is also well-suited for providing astronauts with interactive spatial instructions during tasks such as equipment maintenance or geological sampling. AR has already been deployed to assist astronauts with complex maintenance procedures. For instance, an AR system was used to guide astronauts through treadmill repairs, replacing cumbersome checklists and improving task performance [36]. Other studies have demonstrated AR's ability to reduce both temporal and mental workload during simulated maintenance tasks [35]. For future human space flight missions, similar AR applications could support astronauts in a variety of tasks, from scientific experiments to equipment repair [41].

Here, instructions on medical emergencies during deep space missions have also been proposed in the prior literature [42]. AR systems could therefore offer step-by-step instructions for diagnosing and treating medical conditions when professional assistance is not readily available due to communication delays or radio blackouts. The IPA could provide the AR system with reliable knowledge about procedure steps and relevant multimodal data that can be additionally displayed in the HMD.

*4.3 Human-in-the-Loop Knowledge Graph Updates and Adaptive Head-Up and Head-Mounted Display Systems*

The human-in-the-loop approach proposes that human operators play a central role in ensuring correct system operation and provide manual input if needed [43]. In the context of astronaut assistance this can take place in two ways: 1) HUD and HMD elements can be automatically adjusted based on the learned preferences or requirements of the astronaut and given mission data, such as the current procedure or location. This is done through the update and enrichment of the KG with new information based on experiences made during astronaut training or during EVA/IVA completion. 2) Moreover, manual feedback can be added by the astronaut in case the IPA provides wrong or inaccurate information. To further enhance accuracy, added KG nodes could additionally be checked or edited by mission control or the astronauts after mission completion [44].

The combination of KG updates and RAG-driven responses, along with HUD and HMD adjustments, creates an adaptive, human-centered system that evolves based on astronaut behavior and mission demands.

*4.4 Enhanced Rover Operations Through Head-Up and Head-Mounted Display Integration*

HUDs and HMDs can provide astronauts with live feeds from rover cameras in their field of view, whilst completing EVAs. This system would allow astronauts to visualize key landmarks, potential hazards, and optimal navigation routes before physically encountering them [45].

The data collected by the rovers can be used to enrich the KG, automatically adapting navigation instructions and providing relevant information for both astronauts and rovers. For example, based on the KG, points of interest can be identified, or the rover can be instructed to pre-explore an EVA traverse route.

This improves astronaut safety and operational efficiency by identifying potential hazards, determining the best navigation routes, and identifying suitable geological sampling sites in advance.

*4.4.1 Astronaut-Astronaut Collaboration*

AR can also improve collaboration between astronauts positioned at different locations during missions. By streaming live AR-enhanced footage from helmet cameras, one astronaut can share their perspective with a colleague, increasing the sharing of mental models between both astronauts. This capability is particularly valuable in emergency scenarios, where rapid decision-making is crucial, and multiple team members need to work together to resolve an issue. In such situations, the IPA could serve as an intermediary, analyzing data of the KG and recommending the best course of action based on factors like environmental conditions, astronaut health, and mission objectives. Therefore, multiple astronauts could work more effectively together using AR and IPA technology.

*4.4.2 Astronaut Training*

AR HMD and HUD technology can be utilized not only to assist astronauts during future space missions by overlaying relevant information in real-time but also as a powerful tool for training. It allows for the simulation of training scenarios and procedural malfunctions, enabling astronauts to practice in diverse and challenging environments. This includes simulating hazardous situations, such as fires or smoke, or medical emergencies in a safe, location-independent setting [46].

The IPA can adjust the level of guidance it provides, enabling astronauts to start with a higher degree of support during initial training sessions. As they become more proficient, the IPA can gradually reduce assistance, helping the astronaut transition smoothly to performing procedures with minimal support in real-world situations. This synergy between AR technology and adaptive AI has been shown to be advantageous, for instance, in the context of military training [47].

## 5. Technical Challenges and Potential Solutions

*5.1 Bandwidth and Latency Constraints*

Deep-space missions are constrained by limited communication bandwidth and significant latency. This poses challenges for transmitting large amounts of data, such as AR feeds or complex AI queries, back to Earth.

To address this, local processing on spacecraft or AR devices can reduce the need for constant communication with Earth. Data prioritization techniques could also help optimize bandwidth usage, ensuring that critical information is transmitted first.

*5.2 System Reliability and Redundancy*

System reliability is crucial for the success of long-duration missions. AI systems must be robust, and backup systems should be available to ensure continuous operation in case of failures.

For instance, backups of IPAs with access to archived KGs could be deployed, allowing one to take over in the event of a failure in another. Backup AR systems should also be available to ensure that astronauts can still receive critical information.

*5.3 Head-Up and Head-Mounted Display Technology Under Space Conditions*

A significant technical challenge for HMD and HUD technology in space exploration, particularly in lunar environments, is managing the extreme contrast between bright and dark regions due to the absence of atmospheric light scattering. Astronauts will have to face both blinding sun reflections and pitch-black shadows [48]. Adding to this issue, dust accumulation on astronauts' visors can severely obstruct visibility.

To overcome these challenges, HMD and HUD systems should incorporate advanced image processing technologies that can dynamically adjust brightness, contrast, and focus in real time [49]. Additionally, implementing real-time filtering algorithms to reduce glare and manage shadows will be crucial. Using anti-static coatings, such as NASA's Electrodynamic Dust Shield (EDS) [50], or self-cleaning visors can help minimize dust accumulation, ensuring that the displays remain clear and operational [51, 52].

Additionally, ensuring reliable tracking of 3D-generated content in such conditions is crucial. Although inertial measurement units (IMUs) can still provide important rotational data through gyroscopes, their ability to measure vertical acceleration is impaired by the Moon's or Mars's low gravity or the microgravity environment in space, making the development of complementary tracking methods necessary. In addition, the relatively featureless terrain of the Moon or Mars in certain regions can limit the effectiveness of visual-based tracking, which typically relies on distinct environmental characteristics for accurate localization. To our best knowledge, current HUD prototypes integrated into spacesuits [26] only feature the display of 2D content with a fixed perspective and no spatial information. Given the advantages of spatial AR content proposed in prior sections, a visor-integrated AR HMD system needs to be developed that can change the content displayed in the HMD when the astronaut changes their perspective.

## 6. Future Work and Applications
*6.0.1 CORE*

Recently, our team proposed the integration of AR-based instructions with an IPA based on LLMs and KGs to intuitively display procedural steps during checklist completion.

The project, officially known as the Checklist Organizer for Research and Exploration (CORE) [17], is centered around the design of IPAs in support of astronauts in space. This system is aimed at improving the management of procedures, particularly for missions onboard the ISS, the Lunar Gateway, and future deep-space explorations.

CORE addresses key limitations seen in earlier IPA systems like CIMON and Amazon Alexa, which were rule-based, lacked flexibility, and required constant online connectivity. Unlike these earlier systems, CORE leverages a combination of KG, RAG, and GPT. These technologies ensure that astronauts receive accurate, real-time, and contextually relevant information during spaceflight procedures. The system is also designed to operate offline, an essential feature for space environments where communication with Earth is delayed or unavailable. CORE could be integrated into METIS and connected to the four agents. This would enable CORE to access information from the monitoring agent, which may include multimodal data as described in Section 3.1, or visualize and report anomalies. The results of the reasoning, planning, and commanding agents could be reviewed, confirmed, or - if needed - edited using natural language. This process could become increasingly automated, as suggested in previous works [2].

One of CORE's key features is its AR integration. By utilizing AR elements in a HMD, astronauts can access procedure steps and relevant linked graph information hands-free, reducing their cognitive load and enhancing their spatial understanding. For example, 3D representations of task steps are displayed over real equipment, improving the efficiency of complex tasks like equipment maintenance or emergency procedures.

*6.1 Expansion to Eurocom and Mission Control*

The technologies discussed above have broader applications beyond supporting astronauts in space. As demonstrated by the METIS project [2], IPAs could support mission control procedures by, for example, automating data monitoring activities. As described in CORE, such a system could further enhance decision-making by offering access to relevant data, such as through intuitive visualizations, allow-ing mission control staff to take more informed and timely decisions.

For example, IPAs could monitor spacecraft systems and alert mission control to any anomalies. AR would then provide mission control staff with a 3D visualization of the spacecraft, allowing them to better understand its current state.

*6.2 Applications Beyond Space Exploration*

The combination of AI, KGs, and AR also has potential applications in fields such as healthcare, industrial maintenance, and military operations. In healthcare, AR could help surgeons by overlaying critical information on the patient [53], while IPAs provide real-time guidance based on medical knowledge.

Similarly, in industrial settings, maintenance workers could use AR to receive real-time instructions and feedback while working on complex machinery assembly and maintenance tasks [54]. AI systems could ensure that they are following the correct procedures and provide additional information when needed.

## 7. Conclusion

As humanity ventures further into space, the need for autonomous, reliable systems to support astronauts becomes increasingly important. By integrating IPAs with KGs, RAG and AR, we can provide astronauts with the tools they need to operate independently from mission control. The inclusion of rover integration further enhances the capabilities of these systems, allowing astronauts to perform tasks more efficiently and safely. Although challenges remain, the potential benefits of these technologies make them a critical area for future research and development.

75$^{th}$ International Astronautical Congress (IAC), Milan, Italy, 14-18 October 2024.[32] N. A. Stanton and J. Edworthy, "Auditory warnings and displays: An overview," in *Human factors in auditory warnings*. Routledge, 2019, pp. 3–30.

[33] C. D. Wickens and J. Long, "Object versus space-based models of visual attention: Implications for the design of head-up displays," *Journal of Experimental Psychology: Applied*, vol. 1, no. 3, p. 179, 1995.

[34] R. S. McCann and D. C. Foyle, "Scene-linked symbology to improve situation awareness," in *AGARD Conference Proceedings AGARD CP*, Citeseer, 1996, pp. 16–16.

[35] A. M. Braly, B. Nuernberger, and S. Y. Kim, "Augmented reality improves procedural work on an international space station science instrument," *Human factors*, vol. 61, no. 6, pp. 866–878, 2019, Publisher: SAGE Publications Sage CA: Los Angeles, CA.

[36] V. Byrne, J. Mauldin, and B. Munson, "Treadmill 2 augmented reality (t2 ar) iss flight demonstration," NASA Johnson Space Center, Tech. Rep., 2019. [Online]. Available: https://ntrs.nasa.gov/api/citations/20190028716/downloads/20190028716.pdf

[37] N. J. Anastas, "Augmented reality navigation system for human traversal of rough terrain," Ph.D. dissertation, Massachusetts Institute of Technology, 2020.

[38] Y. E. Cogurcu, J. A. Douthwaite, and S. Maddock, "Augmented reality for safety zones in human-robot collaboration," in *Computer Graphics & Visual Computing (CGVC) 2022*, Eurographics Digital Library, 2022.

[39] C.-C. Mao and F.-Y. Chen, "Augmented reality and 3-d visualization effects to enhance battlefield situational awareness," in *International Conference on Human Interaction and Emerging Technologies*, Springer, 2019, pp. 303–309.

[40] F. A. Rometsch, A. E. Casini, A. Drepper, A. Cowley, J. C. de Winter, and J. Guo, "Design and evaluation of an augmented reality tool for future human space exploration aided by an internet of things architecture," *Advances in Space Research*, vol. 70, no. 8, pp. 2145–2166, 2022.

[41] M. Landgraf, K. Goodliff, C. Esty, and M. Picard, "Lunar surface concept of operations for the global exploration roadmap," NASA Technical Reports Server (NTRS), Tech. Rep., 2021. [Online]. Available: https://ntrs.nasa.gov/api/citations/20210021848/downloads/20210021848.pdf

[42] M. Ebnali *et al.*, "Ar-coach: Using augmented reality (ar) for real-time clinical guidance during medical emergencies on deep space exploration missions," *Healthcare and Medical Devices*, vol. 51, p. 67, 2022.

[43] W. Li, D. Sadigh, S. S. Sastry, and S. A. Seshia, "Synthesis for human-in-the-loop control systems," in *Tools and Algorithms for the Construction and Analysis of Systems: 20th International Conference, TACAS 2014, Held as Part of the European Joint Conferences on Theory and Practice of Software, ETAPS 2014, Grenoble, France, April 5-13, 2014. Proceedings 20*, Springer, 2014, pp. 470–484.

[44] E. Manzoor, J. Tong, S. Vijayaraghavan, and R. Li, "Expanding knowledge graphs with humans in the loop," *arXiv preprint arXiv:2212.05189*, 2022.

[45] H. Ji, S. Li, J. Chen, and S. Zhou, "On-site human-robot collaboration for lunar exploration based on shared mixed reality," *Multimedia Tools and Applications*, vol. 83, no. 6, pp. 18 235–18 260, 2024.

[46] B. Burian *et al.*, "Using extended reality (xr) for medical training and real-time clinical support during deep space missions," *Applied Ergonomics*, vol. 106, p. 103 902, 2023.

[47] K. M. Stanney *et al.*, "Performance gains from adaptive extended reality training fueled by artificial intelligence," *The Journal of Defense Modeling and Simulation*, vol. 19, no. 2, pp. 195–218, 2022.

[48] C. H. Null, M. K. Kaiser, T. E. Wolters, J. J. Marquez, A. M. Cooter, and H. C. Dischinger Jr, "Identification of risks to eva created by ambient lighting conditions at the lunar south pole," in *12th International Association for the Advancement of Space Safety (IAASS) Conference*, 2023.

[49] E. Ahn, S. Lee, and G. J. Kim, "Real-time adjustment of contrast saliency for improved information visibility in mobile augmented reality," *Virtual Reality*, vol. 22, pp. 245–262, 2018.
IAC–24–B3,7,12,x87709     Page 11 of 12

75th International Astronautical Congress (IAC), Milan, Italy, 14-18 October 2024.